\documentclass[12pt, a4paper]{memoir} 
\usepackage[french,english]{babel}

\usepackage [vscale=0.76,includehead]{geometry}                

\usepackage{lipsum}
\usepackage{graphicx}
\usepackage{amsmath}
\usepackage{fullpage}
\usepackage{mathptmx} 
\usepackage{helvet} 
\usepackage{relsize}
\usepackage[T1]{fontenc}
\usepackage{tikz}
\usepackage{booktabs}
\usepackage{textcomp}
\usepackage{multirow}
\usepackage{pgfplots}
\usepackage{url}
\usepackage{footnote}
\usepackage{amssymb}
\usepackage{accents}
\usetikzlibrary{arrows,shapes,positioning,shadows,trees}
\makesavenoteenv{tabular}
\makesavenoteenv{table}

\headstyles{komalike}
\nouppercaseheads
\chapterstyle{dash}
\makeevenhead{headings}{\sffamily\thepage}{}{\sffamily\leftmark} 
\makeoddhead{headings}{\sffamily\rightmark}{}{\sffamily\thepage}
\makeoddfoot{plain}{}{}{} 
\makeheadrule{headings}{\textwidth}{\normalrulethickness}

\setsecnumdepth{subsection}

\pretitle{\HUGE\sffamily \bfseries\begin{center}} 
\posttitle{\end{center}}
\preauthor{\LARGE  \sffamily \bfseries\begin{center}}
\postauthor{\par\end{center}}
\newcommand{\jury}[1]{%
\gdef\juryB{#1}} 
\newcommand{\juryB}{} 
\newcommand{\session}[1]{%
\gdef\sessionB{#1}} 
\newcommand{\sessionB}{} 
\newcommand{\option}[1]{%
\gdef\optionB{#1}} 
\newcommand{\optionB} {}

\option{Data Science} 
\title{Résumé abstractif à partir d'une transcription audio} 
\author{Ilia Derkach}
\date{June 11, 2024} 
\jury{
Research project performed at LIG, APTIKAL team \\\medskip
Under the supervision of:\\
Prof. Massih-Reza Amini \\\medskip
Defended before a jury composed of:\\
Prof. Sana Louhichi\\
Prof. Massih-Reza Amini \\
}
\session{June \hfill 2024}
\begin{document}
\selectlanguage{English} 
\frontmatter
\begin{titlingpage}
\maketitle
\end{titlingpage}

\setlength{\parskip}{-1pt plus 1pt}

\renewcommand{\abstracttextfont}{\normalfont}
\newcolumntype{P}[1]{>{\centering\arraybackslash}p{#1}}
\abstractintoc
\begin{abstract} 
 Currently, large language models are gaining popularity, their achievements are used in many areas, ranging from text translation to generating answers to queries.  However, the main problem with these new machine learning algorithms is that training such models requires large computing resources that only large IT companies have. To avoid this problem, a number of methods (LoRA, quantization) have been proposed so that existing models can be effectively fine-tuned for specific tasks. In this paper, we propose an E2E (end to end) audio summarization model using these techniques. In addition, this paper examines the effectiveness of these approaches to the problem under consideration and draws conclusions about the  applicability of these methods.
\end{abstract}
\abstractintoc

\renewcommand\abstractname{Acknowledgement}
\begin{abstract}
I would like to thank my advisor Prof. Massih-Reza Amini for his
mental support, managing work and our discussions related to
the topic of the master thesis. It was an essential part to finish my master thesis.

I am very thankful for the expert support and discussions with the members of APTIKAL team whose deep knowledge and background helped me with some of the parts of the work. Last but not least I would like to thank all people who organised and participated in the team's scientific seminar, which helped a lot not only find interesting approaches for the research as well as find out something new in closed subjects.
I also place on record, my sense of gratitude to one and all, who directly
or indirectly, have lent their hand in this work.

This work has been partially supported by MIAI @ Grenoble Alpes, (ANR-19-P3IA-0003) funded by the French program Investissement
d’avenir. I am very thankful that they chose me as a candidate for the
scholarship.
\end{abstract}

\cleardoublepage

\tableofcontents* 
\normalsize

\mainmatter
\SingleSpace
\chapter{Introduction}
In the modern world, various machine learning models are widely used in a large number of areas of our lives. It has become commonplace to use such modes not only to predict classes or numerical values (as, for example, in a regression problem), but also to predict and generate more complex objects such as text \cite{1}, audio \cite{2}, or even video \cite{3}. The task of audio summarization can be distinguished. This assumes to create a model which is able to generate a brief description of what was said based on recorded speech. If in the past even the task of summarizing text looked quite difficult and there were no models showing a result at least somewhat comparable to human evaluation, then with the advent of large language models, the solution of such a task became achievable.

It is important to note that such a task is not interesting only in theoretical terms. Such model can be implemented in many modern applications that are used by millions of people every day. For example, this type of model could be implemented to generate automatic video descriptions on popular video platforms, or the generated summary could be used as push notifications for audio messages in common used messengers. Such a solution could also save a lot of time for employees of companies who, due to their work responsibilities, spend a lot of time in online meetings. At the end of each session, the results, decisions and plans formulated to the end of the working meeting could be generated by audio summarization model.

Although it was noted above that the appearance of LLM gave a great breakthrough in the study of many tasks, in particular for the task of audio summarization, however, such models have their own drawback. Due to the presence of a large number of weights and a rather complex architecture, the complete training of such models is time-consuming and computationally intensive. For example, Chat-GPT \cite{4}, the most widely known LLM, has 175 billion parameters and 570 GB of training data. Although the training time for this model is officially unknown, but for comparison Kandinsky model \cite{5} (12 billion parameters) could be taken, whose training took 20,352 GPU-V100 days. Unfortunately, such limitations mean that the development of such models “from scratch” by own capabilities or by the capacities of the research laboratory group is simply impossible at the moment. At the same time, in practice it turned out that models already trained by large companies have sufficient potential to solve more specialized tasks. Due to the large number of weights and the large volume of the dataset being trained, such models assimilate the structure of the language well, therefore, their fine-tuning requires computing power of a much smaller size when using various tactics of retraining.

Speech-to-text (S2T) summarization typically employs a cascade approach \cite{6, 7, 8}, where an automatic speech recognition (ASR) \cite{9, 10} model generates transcripts, followed by a text-to-text (T2T) summarization model \cite{11, 12} that produces summaries. Advances in deep learning, particularly attention-based architectures and self-supervised pre-training, have significantly enhanced the performance of both models. Cascade abstractive systems using these advanced components perform well on dialogue summarization tasks when trained on unpaired data. However, the transcripts produced by the ASR model may contain errors, prompting the development and use of methods such as DL networks or language models to improve robustness and mitigate the impact of these errors.

One major limitation of cascade systems for S2T (Speech to Text) summarization is their inability to utilize non-verbal and acoustic information, such as intonation, pauses, and speaker emotions, which could provide valuable context for more accurate summarization. To address this issue, end-to-end (E2E) modeling has been proposed. E2E systems bypass the intermediate speech recognition step and instead jointly optimize both acoustic and language models in a unified framework. This approach can potentially capture richer information from the audio signal directly, leading to more accurate and contextually aware summaries.

However, E2E modeling presents its own set of challenges. It requires large amounts of paired audio/summary data for effective training, which is often scarce, especially in specific domains like broadcast news. The limited availability of large, publicly accessible corpora necessitates the development of techniques to leverage external data sources. For instance, transfer learning and data augmentation strategies can be employed to enhance model performance. Additionally, unsupervised and semi-supervised learning approaches can help make better use of unpaired or partially paired datasets, further addressing the data scarcity issue.

This work proposes E2E model for S2T abstractive
summarization. The model uses fine-tuned ASR and T2T abstractive summarizer on a descriptive videos dataset using such methods as LoRA \cite{13, 14, 15} and quantization \cite{17, 18}. The E2E system follows the encoder-decoder paradigm and utilizes speech features extracted using a speech representation of wave forms provided in the dataset. As mentioned above, due to the specifics of the model, a text summarization model and a speech recognition model were allowed. Metrics computed for these models were compared with other classical results in these spheres, which allowed us to draw conclusions about the applicability of the methods of fine-tuning described above.

This paper is organized as follows: in section 2, I present the description of effective fine-tuning LLMs such as LoRA and quantization, which were used in the experiments; in section 3, I am describing the dataset on which the training was performed and the architecture of T2T (text to text) abstractive summarizer is described, as well as the presentation of the results and comparison with existing models; in section 4, the same thing is written for the ASR model; in section 5, I present how the trained models can be combined into a single E2E speech-to-text model, and also describe plans for the future research. 
\chapter{PEFT: State-of-the-art Parameter-Efficient Fine-Tuning}
In the rapidly evolving field of machine learning, the quest for more efficient and scalable models has led to significant innovations in model fine-tuning techniques. Parameter-Efficient Fine-Tuning (PEFT) has emerged as a pivotal approach, offering the ability to adapt large pre-trained models to specific tasks with a fraction of the computational cost and resources. This chapter delves into the PEFT library \cite{19}, focusing particularly on Low-Rank Adaptation (LoRA) \cite{13, 14, 15} and quantization methods \cite{17, 18}, which I have used for fine-tuning models for my specific tasks.

\section{Low-Rank Adaptation (LoRA)}

To demonstrate the operation of this mechanism, consider an ordinary linear layer represented as follows
\[
y = Wx,
\]
where $x$ is the input of the layer, and $W$ is the matrix of weights. In the process of further training the model, it is necessary to slightly change the operation of this layer by adjusting the weights by $\Delta W$ (which are usually searched for by the gradient descent), so that the new output would be as follows
\[
y' = W'x = (W + \Delta W)x = Wx + \Delta Wx
\]
As can be noted, the new $y$ value differs from the old one by $\Delta W$, which can be interpreted as the result of the work of a single, separate, fully connected layer. This interpretation is demonstrated in the Figure \ref{fig:lora} below.
\begin{figure}[!ht]
    \centering
    \includegraphics[width=0.4\textwidth]{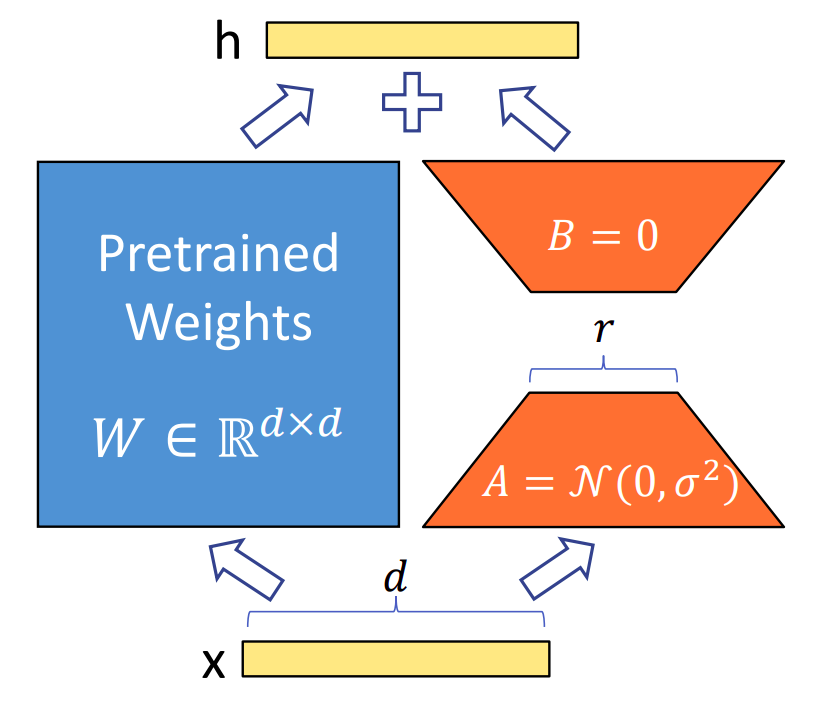}
    \caption{Schematic representation of LoRA. Weights $W$ are frozen, while $\Delta W = A\cdot B$}
    \label{fig:lora}
\end{figure}
Thus, the weights of the matrix $W$ can be fixed, and instead the weights of $\Delta W$ can be trained. This can be interpreted as training a model that predicts the difference between the result of a conventional model and a pre-trained one. This is vaguely reminiscent of gradient boosting, where each subsequent decision tree is trained to correct the errors of the previous model.

However, a legitimate question may arise why this transformation is an optimization if the number of weights in $W$ and $\Delta W$ is the same. Just to reduce the number of weights, matrix decomposition into the product of two matrices of lower rank is used, as mentioned in the name of this method. Speaking in more detail, the matrix $\Delta W$ of size $n\times k$ is represented as $A\cdot B$, where $A$ and $B$ have sizes $n\times r$ and $r\times k$, where $r$ is a small constant. In the original article \cite{13}, it was shown that it is enough to choose a fairly small $r$ with rank of 8-128, in which case the number of trained parameters will be less in
\[
\frac{n \cdot k}{(n + k) \cdot r} \approx 10^2, \text{in general.}
\]
The question may arise whether the model will lose generalizing complexity if we choose the parameter $r$ small enough. However, the authors of the article \cite{13} demonstrated that in real life the "inner rank" of large text models is very low, and most of the parameters do not affect a lot. 

When initializing the model, matrix B is set randomly (for example, from $\mathcal{N}(0, \sigma^2)$, and matrix $A$ is initialized with zeros, so that initially $\Delta W = 0$.

These layers can be perceived as adapters to linear layers. If we talk about the transformer architecture (which, as will be shown later, includes the E2E and ASR models being trained), then usually such layers are applied to self-attention matrix weight. In detail, let self-attention module be represented as
\[
Self{\text -}Attention(Q, K, V) = softmax\left(\frac{QK^T}{\sqrt{d_k}} \right)V,
\]
where $d_k$ is the dimension of the keys and queries (one of the dimension of matrices $Q$ and $K$) and matrices $Q$ (queries), $K$(keys), and $V$ (values) are linear projections of input $X$:
\[
Q = XW^Q \;\;\;\;\;\;\;\;\;\; K = XW^K \;\;\;\;\;\;\;\;\;\;  V = XW^V,
\]
where $W^Q$, $W^K$, $W^V$ are learned projection matrices. Then the LoRA method will be applied to matrices $W^Q$, $W^K$, $W^V$.

Although this method is quite effective in terms of memory usage, as well as learning time, it has a significant drawback. Transformers have many attention layers to which LoRA is applied. However, the rank $r$ of the decomposition is the same for all adapters, which is certainly inefficient: it is clear that some matrices contribute more and have a high rank, while others have minor. This drawback has been fixed in an improved version of the algorithm called AdaLoRA \cite{14}, which will be discussed in the next section.

\section{Adaptive budget allocation for parameter-efficient fine-tuning (AdaLoRA)}

To correct the inefficiency described in the section before, method AdaLoRA \cite{14} was proposed that adaptively allocates the parameter budget among weight matrices based on their importance scores. AdaLoRA parameterizes the adapters using a form of their singular value decomposition (SVD) \cite{20}. This innovative approach enables effective pruning of the singular values of less important updates, thereby reducing their parameter budget while avoiding the need for intensive exact SVD computations.

Getting into the details, let the LoRA adapter $\Delta W$ have the following singular value decomposition:
\[
\Delta W = P\Lambda Q, 
\]
where $P \in \mathbb{R}^{d_1 \times r}$ and $Q \in \mathbb{R}^{r \times d_2}$ represent the left/right singular vectors of $\Delta W$ and $\Lambda \in \mathbb{R}^{r \times r}$ is a diagonal matrix with singular values, which corresponds to left/right singular vectors in the matrices $P$ and $Q$ respectively. While the previous method trains the matrices $A$ and $B$ of arbitrary dimension $r$, in the current approach the matrices $P$, $Q$, $\Lambda$ from SVD are trained.

It is important to understand that in addition to the change in the trained matrices, the training process also changes slightly. If earlier the learning error was some training cost $\mathcal{C}(P, \Lambda, Q)$, which demonstrates the degree of difference between the output of the current model and the target value, then in the current algorithm a regularization term is added to the  training cost, which is designed to maintain the orthogonality of the matrices $P$ and $Q$. Formally, the updated loss function for each layer to which AdaLoRAA is applied looks like
\[
\mathcal{L}(P, \Lambda, Q) = \mathcal{C}(P, \Lambda, Q) + \gamma \begin{Vmatrix}P^TP - I\end{Vmatrix}_F^2 + \gamma \begin{Vmatrix}QQ^T - I\end{Vmatrix}_F^2
\]
In addition, the process of updating the trained weights is also changing. If nothing changes for the $P$ and $Q$ matrices, and they continue to be updated using stochastic gradient descent over the updated loss function, then the $\Lambda$ update changes in order to change the rank of the product matrix $\Delta W$. Specifically, for the matrix $\Lambda$ the stochastic gradient is made as follows (only for diagonal elements, the rest remain equal 0):
\[
\tilde{\Lambda}^t = \Lambda^t - \eta \nabla \mathcal{L}(P^t, \Lambda^t, Q^t), 
\]
where $\eta > 0$ is gradient descent learning rate. However, after that the singular values in $\tilde{\Lambda}^t$ are pruned based on importance score $S^t$:
\[
\Lambda^{t + 1}_{i, i} = \begin{cases}
  \tilde{\Lambda}^t_{i, i} & \text{if $S^t_i$ in the top-$b^t$ of $S_t$} \\    
  0 & \text{otherwise},   
\end{cases} 
\]
where $S^t$ contains the importance score of all singular values and its' left/right singular vectors and $b^t$ is the budget of remaining singular values at the $t$-th step. $b^t$ should be considered as decreasing function depends on $t$, allowing us efficiently eliminate the proportion of unimportant singular numbers. At the same time, in the simplest example $S^t$ may only rely on singular values and be equal to it's absolute value: $S^t_i = |\tilde{\lambda}^t_i|$ (a large singular value is more important to preserve in pruning). In reality it was shown that is is worth to choose more complicated function depending both the singular values and corresponding singular vectors. 

\section{Quantization}

Quantization is the process of converting values from a representation with a large amount of information (usually a continuous set) to a more compact representation, usually from a discrete set. A good example of quantization is signal sampling, when each value of a continuous signal is assigned a value from a predefined discrete set. In the context of neural networks, quantization means moving from a data type with a large number of bits, such as \texttt{float32}, to a type with a smaller number, such as \texttt{int8}. This approach allows to significantly reduce the weight of the model without greatly impairing its quality. This may be critical, for example, for models running directly on mobile devices. Talking about LLM state-of-the-art models, it is impossible to save models directly on the phone inside the mobile application, because the amount of memory of the model does not allow this. If the application itself does not allow the developers to work with the model on the server, then without quantization it will not be possible to save it on the phone. In addition, quantized models require less computing resources and run faster. Further it will be discussed in more detail on how this transformation works for the weights of neural networks.

There are different approaches to quantification. However, only the two most popular will be considered as they are used in the current work. The first, simplest and most intuitive is the linear transformation. In general terms, this means that the original range of values $[R_{\min}; R_{\max}]$ transforms into the quantized range $[Q_{\min}; Q_{\max}]$ by affine transformation. It can be divided into two types: asymmetric and symmetric quantization. Let's consider the asymmetric transformation at the beginning. 

Let's denote $S$ and $Z$ as quantization constants, in other words parameters that are calculated in the process. Define $S$ as follows 
\[
S = \frac{R_{\max} - R_{\min}}{Q_{\max} - Q_{\min}}
\]
The constant $S$ named scale and stored in the initial type. The constant $Z$ named zero-point and define the point where initial zero value will be transformed. Mathematical expression for $Z$ is
\[
Z = \left[Q_{\min} - \frac{R_{\min}}{S} \right]
\]
It is very important for neural networks to accurately represent zero, so this constant is determined. Rounding in the definition above can be done in different ways: either down or up, or even to the nearest integer. $Z$ in opposite to $S$ is often stored in a quantized type. The quantization and dequantization functions are as follows:
\[
X_q = \left[\frac{X}{S} + Z \right] \;\;\;\;\;\;\; \text{and} \;\;\;\;\;\;\; X = S(X_q - Z).
\]
Asymmetric quantization is well suited for non-symmetric distributions, for example, for ReLU output.

Symmetric quantization works similarly, with the difference that the zero-point is zero, which ensures that the values are symmetric relative to zero. Additionally, the boundaries of the quantized range are defined as the maximum modulo the quantized value $|R_{\max}|$. Finally, in order for the type to be symmetric, it requires to discard one value in the quantized data type. For example, the range \texttt{unsigned int8}: [-128, 127] will turn into [-127, 127]. The above defined constants are defined as follows
\[
S = \frac{|R_{\max}|}{2^{N - 1} - 1} \;\;\;\;\;\;\; \text{and} \;\;\;\;\;\;\; Z = 0,
\]
where $N$ is a number of bits in quantized type. The quantization and dequantization are
\[
X_q = \left[\frac{X}{S} \right] \;\;\;\;\;\;\; \text{and} \;\;\;\;\;\;\; X = SX_q
\]
The second approach in the transformation of weights during quantization is based on the fact that the weights are not distributed linerarly over the entire range, but in the form of a normal distribution. Futher it will be considered that all the weights are in range $[-1, 1]$. In this regard, it makes sense to give more values from the quantized type to the weights from the middle of the range, and less values to the weights from the edges. An example of this approach is NF4, the construction of a 4-bit data type that was used in experiments in this paper.

There are 4 bits, it means that only 16 values can be stored. Two values should represent -1 and 1, so only 14 remains. They are used to represent some quantiles $\mathcal{N}(0, \sigma^2)$ in the range [-1, 1]. After that, each weight can be correlated with the nearest value from the previously calculated quantiles. However, the described scheme has a drawback, it does not have an exact representation for zero. As it was mentioned before the exact value of zero is critically important for neural networks, for example, to be able to do padding and other zero-valued elements with no error, so the authors offer an elegant solution. Firstly, the range [-1, 1] is divided into two parts, positive and negative. Then, 7 quantiles is found in the negative part and 6 quantiles in the positive one. Finally, zero is added to the obtained values.

A neural network can be quantized with different granularity. The worst way is to quantize the entire network at once. In this case, one common constant $S$ will be received for the entire model. The result of such manipulations is likely to be unsatisfactory. It is also possible to quantify the tensors individually. Thus, each tensor will get its own constants, or it is even possible to quantize rows or columns in each tensor. Accordingly, each row or column in this case will have its own constant. It is also possible to divide tensors into small blocks and carry out quantization within blocks. Although quantized vectors should to be stored somewhere optimaly, at the same time the calculations will be more accurate. Making a small summary, it can be claimed that with smaller granularity, the fewer constants need to be stored, and vice versa — the higher the granularity, the closer the results of quantized calculations are to the original ones.

\section{Efficient Finetuning of Quantized LLMs (QLoRA)}

One crucial aspect of Low-Rank Adaptation (LoRA), the method which was discussed before, is its memory requirements during training, specifically concerning the number and size of adapters used. Thanks to LoRA's minimal memory footprint, more adapters can be employed to boost performance without significantly increasing overall memory usage. Thus, a kind of bottle neck is that the main memory consumption during the fine-tuning of large language models comes from activation gradients in the frozen weights of the original model rather than the LoRA parameters. For example, in a 7 billion parameter LLaMA \cite{21} model trained on the FLAN v2 dataset \cite{22} with a batch size of 1, the memory usage is as follows: LoRA weights, typically about 0.2\% of the original model weights, occupy only 26 MB. In contrast, the memory footprint for the LoRA input gradients is 567 MB. Quantization allows the developers to reduce the amount of memory required for calculating gradients in several times.

However, it is not enough to simply quantify the weights of the original model. The fact is that when adding up the results of the LoRA adapters and the model weights, their type must be the same. At the same time, it is inefficient to conduct training in adapters in a quantized form (\texttt{int4} or \texttt{int8}). In this regard, the authors have come up with a new QLoRA \cite{16} approach, which will be discussed further. 

Using the components described above, we define QLoRA for a single linear layer in the quantized base model with a single LoRA adapter as follows
\[
\text{\textbf{Y}}^\text{BF16} = \text{\textbf{X}}^\text{BF16}\text{dequant(\textbf{W}}^\text{NF4}) + \text{\textbf{X}}^\text{BF16}\text{\textbf{A}}^\text{BF16}\text{\textbf{B}}^\text{BF16}, 
\]
where $A$ and $B$ are LoRA adapters and $W$ is quantized matrix of the initial linear layer. For parameter updates only the gradient with respect to the error function $L$ for the adapters weights $\frac{\partial L}{\partial A}$ and $\frac{\partial L}{\partial B}$ are
needed, and not for 4-bit weights $\frac{\partial L}{\partial W}$ . However, the calculation of $\frac{\partial L}{\partial A}$ and $\frac{\partial L}{\partial B}$
entails the calculation of $\frac{\partial X}{\partial W}$
which proceeds via equation above with dequantization from storage $\text{\textbf{W}}^\text{NF4}$ to computation data type
$\text{\textbf{W}}^\text{BF16}$ to calculate the derivative $\frac{\partial X}{\partial W}$ in BFloat16 precision.

Instead of NF4 type, there can be other types of quantization, as well as double quantization: a method in which the number of bits decreases gradually, by sequentially applying two quantizations with an increasing block size, according to which quantization is done.
\chapter{Text-to-text summarization}

Text summarization is a crucial task in natural language processing (NLP) that involves generating concise and coherent summaries of larger texts. The goal is to capture the most important information and convey it in a shorter form while retaining the original meaning and context. Text summarization has a wide range of applications and challenges. It is used in news summarization to create concise summaries of news articles, document summarization to summarize long reports, research papers, and legal documents, and dialogue summarization to condense conversations or meeting transcripts. However, it also presents several challenges, such as maintaining coherence to ensure that the summary is logically coherent and readable, covering all critical information while being concise, handling ambiguities to ensure the generated summary accurately reflects the original text, and creating effective metrics to evaluate the quality and accuracy of summaries. There are two main approaches to text summarization: extractive and abstractive summarization.

Extractive summarization \cite{23, 24, 38, 25} works by selecting and extracting key sentences, phrases, or sections directly from the original text. This approach identifies the most important parts of the text based on various criteria such as sentence importance, word frequency, and positional significance. The selected content is then combined to form the summary. It relies on selecting existing segments of the text, making the generated summaries more straightforward and easier to generate because they are composed of actual sentences from the original text. Since it uses the exact words and phrases from the source, it avoids issues like rephrasing errors or loss of factual accuracy. Techniques used in extractive summarization include frequency-based methods \cite{26} that identify and extract sentences with the highest frequency of important words, graph-based methods that use algorithms like PageRank \cite{27} to score and rank sentences based on their relationships and importance within the text, and machine learning approaches that employ supervised or unsupervised learning techniques to identify and select key sentences.

On the other hand, abstractive summarization \cite{7, 8, 28} involves generating new sentences that capture the essence of the original text. This method goes beyond merely selecting sentences and instead paraphrases and synthesizes information to create a coherent and concise summary. Abstractive summarization involves generating new text that may not directly appear in the original document. It is more complex than extractive summarization, as it requires understanding and rephrasing the content. This approach can produce more human-like, readable summaries by rephrasing and condensing information in a way that might be more natural and coherent. Techniques used in abstractive summarization include Seq2Seq models that utilize encoder-decoder architectures where the encoder processes the input text and the decoder generates the summary, Transformer models that leverage models like BERT \cite{29} or GPT to understand the context and generate summaries, and reinforcement learning that can be used to fine-tune models by rewarding summaries that are coherent and accurate.

\section{MBart}

The MBart model \cite{7}, or Multilingual BART, is a powerful transformer-based sequence-to-sequence model designed for multilingual machine translation and text generation tasks. Developed by Facebook AI, MBart extends the capabilities of the original BART (Bidirectional and Auto-Regressive Transformers) model to support multiple languages, enabling it to handle a wide range of language pairs and multilingual scenarios with high efficiency and accuracy.

MBart leverages a pre-training and fine-tuning approach. During the pre-training phase, the model is trained on a large corpus of multilingual data using a denoising auto-encoder objective. This involves corrupting the input text by masking spans of tokens and then training the model to reconstruct the original text. This process allows MBart to learn robust language representations and understand the syntactic and semantic nuances of different languages.

One of the key features of MBart is its ability to perform zero-shot translation, which means it can translate between language pairs it has never seen during training. This is achieved by using language-specific tokens that help the model identify the source and target languages, allowing it to generalize across languages effectively. For instance, when translating from English to French, special tokens indicating the source language (English) and the target language (French) are added to the input sequence, guiding the model to generate the appropriate translation.

The architecture of MBart consists of an encoder-decoder framework, where both the encoder and decoder are composed of multiple layers of transformers. The encoder processes the input text and generates a series of contextual embeddings, while the decoder uses these embeddings to produce the translated or generated text, one token at a time. This design allows MBart to capture long-range dependencies and generate coherent and contextually appropriate outputs.

MBart supports a wide array of languages, making it versatile for multilingual applications. It is particularly effective in low-resource language scenarios, where it can leverage shared representations from high-resource languages to improve translation quality. The model's ability to handle diverse languages and its pre-training on extensive multilingual data make it a robust tool for tasks like document translation, multilingual text generation, and cross-lingual understanding.

\section{T5}

The T5 model \cite{8}, also known as the Text-to-Text Transfer Transformer, is a versatile and powerful transformer-based model developed by Google Research. T5 is designed to handle a wide range of natural language processing (NLP) tasks by framing them all as text-to-text problems. This means that both the input and output are always treated as text strings, which simplifies the model architecture and training process.

At the core of T5 is the idea that a single model can perform various tasks if those tasks are represented in a consistent format. For instance, tasks such as translation, summarization, question answering, and classification are all converted into text-to-text formats. For example, translating a sentence from English to French is formatted as "translate English to French: [input sentence]", while summarizing a paragraph is formatted as "summarize: [input text]".

T5’s architecture is based on the transformer model, specifically leveraging the encoder-decoder structure. The encoder processes the input text and generates a series of contextual embeddings, which the decoder then uses to produce the desired output text. This design allows T5 to capture complex dependencies within the text and generate high-quality, contextually relevant outputs.

A significant strength of T5 is its extensive pre-training on a massive and diverse dataset called the C4 (Colossal Clean Crawled Corpus) \cite{30}. This dataset comprises a vast amount of web pages, providing the model with a broad understanding of language. During pre-training, T5 is exposed to various tasks, learning to predict missing tokens and understand different linguistic structures. This extensive pre-training equips T5 with a strong general understanding, which it can apply to a wide array of downstream tasks with fine-tuning.

In terms of performance, T5 has achieved state-of-the-art results on numerous NLP benchmarks, including the GLUE, SuperGLUE, and SQuAD datasets. Its ability to handle diverse tasks with a unified approach simplifies the deployment of NLP systems and makes it easier to adapt to new tasks by simply changing the input prefix and fine-tuning on the relevant data.

In summary, the T5 model is a groundbreaking and flexible tool in the field of natural language processing. By treating every task as a text-to-text problem, T5 simplifies the model architecture and training process, allowing it to excel across a wide range of tasks. Its encoder-decoder structure, extensive pre-training on the C4 corpus, and the use of task prefixes make T5 a powerful and versatile model for a multitude of NLP applications.

\section{Summarization metrics}

There are several metrics by which you can measure the accuracy of summarization compared to targeted summarization. It is clear that due to the fact that a good summarization does not necessarily have to match the targeted one precise, the metrics that will be used must somehow take this into account. For example, rearranging some words in the target value should not lead to a zero metric value, while word-by-word comparison will lead to exactly this.

One of the most popular metrics is the ROUGE family \cite{31} of metrics. ROUGE is a set of metrics that compare the overlap between the generated summary and a reference summary. These scores range from 0 to 1, where higher values denote a closer match between the automated summary and the reference. The most common variants are ROUGE-N, ROUGE-L, and ROUGE-S. ROUGE-N measures the overlap of $n$-grams between the generated summary and the reference summary:
\[
\text{ROUGE-N} = \frac{ \sum_{n\text{-gram} \in S} \text{Count}_{\text{match}}(n\text{-gram})}{\sum_{n\text{-gram} \in S} \text{Count}(n\text{-gram})},
\]
where $n$-gram is a contiguous sequence of $n$ items from the text, $\text{Count}_{\text{match}}(n\text{-gram})$ is the number of $n$-grams co-occurring in both the generated and reference summaries and $\text{Count}(n\text{-gram})$ is the total number of $n$-grams in the reference summary. Speaking in more detail, the value of $\text{Count}_{\text{match}}(n\text{-gram})$ is the minimum among the number of $n\text{-gram}$ in the targeted and generated summarization (denoted as $S$ in the formula above). When there is more than one reference summary, then the individual ROUGE scores are calculated per reference and the average is returned.

ROUGE-L uses the longest common subsequence (LCS) between the generated summary and the reference summary:
\[
\text{ROUGE-L} = \frac{LCS(X, Y)}{\text{length}(Y)},
\]
Where $LCS(X,Y)$ is the length of the longest common subsequence between sequences $X$ (generated summary) and $Y$ (reference summary). This metric also varies from 0 to 1.

ROUGE-S, also known as ROUGE-Skip-Bigram, measures the overlap of skip-bigrams between a generated summary and a reference summary. A skip-bigram is any pair of words in their sentence order, allowing for arbitrary gaps. This metric is particularly useful for capturing the presence of important pairs of words that may not appear consecutively but still maintain their order and contextual relevance.  The formula for ROUGE-S is:
\[
\text{ROUGE-S} = \frac{\sum_{\text{skip-bigram} \in S} \text{Count}_{\text{match}}(\text{skip-bigram})}{ \sum_{\text{skip-bigram} \in S} \text{Count}(\text{skip-bigram})},
\]
where skip-bigram represents the pairs of words in their original order, allowing for gaps and $\text{Count}_{\text{match}}()$, $\text{Count}()$ are defined as in the ROUGE-N definition. ROUGE-S captures the preservation of important word pairs in their original order, which helps to maintain the contextual and semantic integrity of the summary. It is particularly useful when the exact ordering of words is less important than capturing the relationships between key terms.

There are other metrics, but in this study we will focus on this family of metrics.

\section{Dataset for text summarization}

At the moment, there are quite a lot of datasets for text summarization. However, for the most part they are either small or highly specialized. This is due to the fact that it is quite difficult to create a summary for a large body of texts, as this requires a lot of human resources. In addition, it is important that the training dataset contains abstract short contents, not extractive ones. Otherwise, the model will not be trained for the type of task and will collect a summary of the full sentences of the original text. 

One of the main sources for abstract text summarization is news. Very often, texts dedicated to various events are accompanied by short summary content to increase the number of views. These texts are written by people themselves, and as a result meet the criteria described above. In addition, the collection of such texts satisfies all copyrights, since news sites make it possible to use texts for non-commercial purposes for quoting and other works with the text. Gazeta dataset proposed by Ilya Gusev \cite{32} was chosen among a class of similar datasets.

Gazeta is a Russian-language news publication that contains both local news and news from around the world. The dataset contains only those summaries which are less than 85 words and more than 15 words, and which texts are less than 1500 words and pairs text-summary with more
than 30\% unigram intersection, and less than 92\% unigram intersection. The last condition is most important, as it can directly affect the final ROUGE metric based on the calculation of $n$-gramms. Thus, ROUGE-2 metric for the dataset is 22.7. In addition, an important characteristic is the average length of texts in a pair: the average length of articles is 767.5, and the average length of a summary is 53.3. Also, as it was mention before, it is quite important to have not-extractive summarizations in the training dataset. To look at this closer, the percentage of new $n$-gramms in the obtained summary will be explore. Below, there is table \ref{tb:1} in which the ROUGE metrics for the dataset itself are calculated.

\begin{table}[htbp!]
\centering
\begin{tabular}{||c | c | c | c ||} 
  \hline
       & Train & Val & Test \\ 
  \hline
  1-gramms & 34.2 & 30.5 & 30.6\\ 
  \hline
  Lemmatized 1-gramms & 21.4 & 17.8 & 17.6 \\ 
  \hline
  2-gramms & 68.6 & 65.0 & 65.5 \\ 
  \hline
  Lemmatized 2-gramms & 61.4 & 58.0 & 58.5 \\ 
  \hline
  3-gramms & 84.5 & 81.5 & 81.9 \\ 
  \hline
\end{tabular}
\caption{Average \% of novel $n$-grams in Gazeta dataset}
\label{tb:1}
\end{table}
In addition to the Russian-language dataset, the study considered an English-language dataset with a different approach to the formation of short content in the training dataset. The main problem of the existing summarization datasets, which consist of news articles is the following: the articles are written by journalists and follow the journalistic style. As professional writers,  journalists usually prioritize and structure their texts by starting with mentioning the most important and attention grabbing elements of a story in the opening paragraphs and later adding details and any background information. This writing style might be the cause why there is a quite low percentage of novel 1-gramms and 2-gramms in Gazeta dataset (the summaries are generally formed by repeating a couple of first sentences of the text or slightly changed) 
usually score higher compared to the existing summarization systems. The authors proposed another way to generate summaries \cite{33}, using WikiHow, a cite with instructions to do something. This dataset contains articles written by ordinary people, not journalists. Therefore, the text are not written by professionals which is more close to the real life.

The WikiHow knowledge base is a collection of online articles that provide step-by-step instructions on a wide range of topics, from arts and entertainment to computers and electronics. Each article features a title beginning with "How to" and includes a brief description. Articles fall into two categories: single-method articles, which outline tasks in sequential steps, and multi-method articles, which offer different approaches for completing a task. Each step in the articles begins with a bold summary line, followed by a more detailed explanation. To generate summaries, only those articles were taken that include several steps. In each article the first sentences in each step were extracted and concatenated in a separate text, which is consider to be a targeted summary, while these extracted sentences are removed from the original text. The final dataset is made of 204004 pairs of articles and their summaries. 

Speaking about the characteristics of this dataset, it is important to say that the average length of the article in the dataset is 579.8, and the average length of the summary is 62.1. The difference between the lengths of the summary with the previous dataset is not significant, so further comparison of the metrics with each other is valid. In addition, a table with percentages of novel $n$-gramms in summaries comparing with the original text, as it was done for Gazeta dataset, provided in table \ref{tb:2}.

\begin{table}[htbp!]
\centering
\begin{tabular}{||c | c ||} 
  \hline
       & Dataset \\ 
  \hline
  1-gramms & 31.1\\ 
  \hline
  Lemmatized 1-gramms & 21.6 \\ 
  \hline
  2-gramms & 78.8 \\ 
  \hline
  Lemmatized 2-gramms & 70.1 \\ 
  \hline
  3-gramms & 93.3 \\ 
  \hline
\end{tabular}
\caption{Average \% of novel $n$-grams in WikiHow dataset}
\label{tb:2}
\end{table}

It is important to note that due to the fact that the dataset was not divided into a training, validation and test datasets, the metrics were calculated for the entire dataset. In addition, it can be seen that for 1-gramms, the metric value has approximately the same value, while the percentage of new 2-gramms and 3-gramms for the wikiHow dataset is significantly higher than for the Gazeta dataset. This difference can be explained precisely by the fact that the articles are written by professional journalists, so there may be more "plagiarism" and extractive summarization in the summary. However, it is worth to mention that such a comparison is not completely correct, since the language of the datasets is different, therefore this statistic (percentage of $n$-gramms) can be heavily language-dependent.

\section{Fine-tuning text summarizarion models}

This section describes the fine-tuning  of the T5 and MBart text summarization models discussed above. After receiving the results, I analyzed the effect of applying LoRA and AdaLoRA (see section 2) to matrices within self-attention layers. Experiments were conducted for different hyperparameters $r$, as a result of which the optimal parameter in terms of accuracy and number of parameters of the trained model was found.

\begin{figure}[htbp!]
\begin{center}
\includegraphics[width=1\columnwidth]{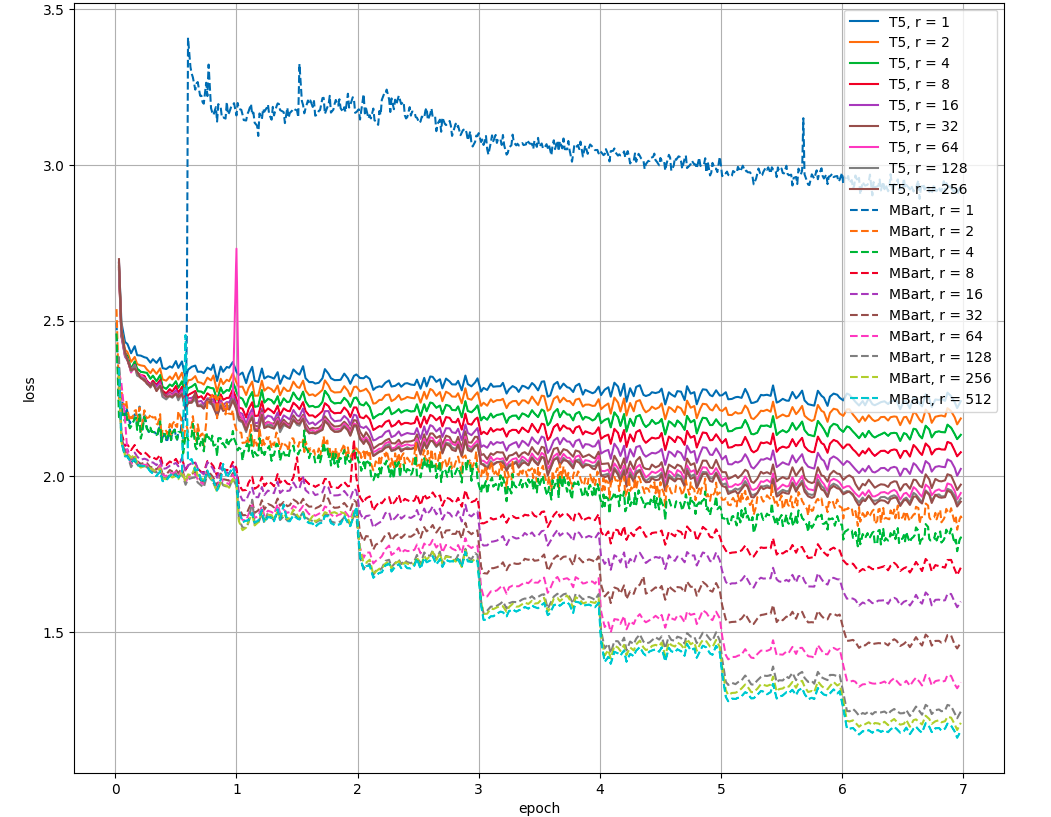}
\end{center}
\caption{Training plot for fine-tuning MBart and T5 models with LoRA with different $r$}
\label{fig:training}
\end{figure}

For example, the training plot for MBart and T5 models is shown in Figure \ref{fig:training}.

The first thing to note is that as the value of $r$ increases, the learning curves decrease. This means that the prediction is very close in terms of the loss function, which is played by cross-entropy on embeddings. It is important to note that a decrease in the loss function does not mean an improvement in the summary given, because the resulting summary may differ greatly in embedding from the target, at the same time, it may be quite accurate. To assess the quality of the model, ROUGE metric is nedded, which more accurately shows the semantic coincidence in two texts. Although, ideally, the most accurate assessment is a human assessment. Unfortunately, it is impossible to assess the quality of the model after each step of the gradient, since it takes an average of 40 minutes to calculate the ROUGE metric on the computing power of the cluster. In this regard, according to the degree of forest decrease, the required number of epochs for training was estimated, after which the ROUGE metric was calculated on the final models (see the tables \ref{tb:3} and \ref{tb:4}). The graph in Figure \ref{fig:training} shows that for most models, 7 epochs are enough for training. Although, at the same time, it is clear that for models with a large hyper parameter, $r$ can be trained further, since the decreasing trend of the loss function is still decreasing, it has been verified that this has no significant effect on the ROUGE metric. Speaking about the number of required epochs for training models, it is important to note that it is not important how many epochs, how many steps (in other words, calculating the gradient on the batch) are made during training. That is why WikiHow dataset training required fewer epochs, five instead of seven, but the completion process was the same in terms of time, because the training sample was larger, and as a result, the total number of steps for Gazeta dataset and WikiHow dataset was approximately the same. Finally, we note that the training time of one epoch ranges from 40 minutes, ending with 100 minutes, depending on the number of parameters (i.e. the value of the parameter $r$) for LoRA adapters.

Now let's focus on the value of the ROUGE metric for the experiments performed. As mentioned above, training similar to the one whose graph is shown above was conducted for two datasets for the T5 and MBart models with full fine-tuned weights, as well as fine-tuning with LoRA and AdaLoRA adapters. The results are presented in the tables \ref{tb:3} and \ref{tb:4}. In the tables for cases where LoRA adapters were used, the best values are highlighted for all possible $r$, so as not to clutter the table. The dependence of metrics on $r$ will be discussed later.

\begin{table}[htbp!]
\centering
\begin{tabular}{||c | c | c | c | c ||} 
  \hline
       & ROUGE-1 & ROUGE-2 & ROUGE-L & ROUGE-S \\ 
  \hline
  Finetuned MBart & 32.1 & 14.2 & \textbf{27.9} & 20.1 \\ 
  \hline
  Finetuned T5 & \textbf{35.3} & 13.1 & 26.5 & 22.4 \\ 
  \hline
  MBart + LoRA (best) & 24.1 & 9.7 & 18.9 & 18.9 \\ 
  \hline
  T5 + LoRA (best) & 24.5 & 9.9 & 19.3 & 19.3 \\ 
  \hline
  MBart + AdaLoRA (best) & 33.4 & \textbf{17.1} & 25.9 & \textbf{23.3} \\ 
  \hline
  T5 + AdaLoRA (best) & 30.1 & 15.7 & 25.4 & 20.3 \\ 
  \hline
\end{tabular}
\caption{ROUGE metrics for models finetuned on Gazeta dataset}
\label{tb:3}
\end{table}

\begin{table}[htbp!]
\centering
\begin{tabular}{||c | c | c | c | c ||} 
  \hline
       & ROUGE-1 & ROUGE-2 & ROUGE-L & ROUGE-S \\ 
  \hline
  Finetuned MBart & \textbf{35.9} & \textbf{13.9} & \textbf{34.8} & 25.4 \\ 
  \hline
  Finetuned T5 & 35.4 & 9.3 & 23.6 & 28.4 \\ 
  \hline
  MBart + LoRA (best) & 27.1 & 10.2 & 22.3 & 20.1 \\ 
  \hline
  T5 + LoRA (best) & 28.5 & 9.1 & 23.3 & 26.3 \\ 
  \hline
  MBart + AdaLoRA (best) & 33.4 & 13.0 & 33.1 & 26.1 \\ 
  \hline
  T5 + AdaLoRA (best) & \textbf{36.1} & 12.1 & 25.4 & \textbf{29.3} \\ 
  \hline
\end{tabular}
\caption{ROUGE metrics for models finetuned on WikiHow dataset}
\label{tb:4}
\end{table}

Among the results obtained, it is important to note that in both cases, the values of the metrics when learning with adapters turned out to be close to what if the model had been fully trained. On average, metrics and quality are slightly worse, but the amount of time spent on training has decreased significantly. So, on average, it takes about 1 hour per epoch to train a model with LoRA adapters, while it takes an average of 34 hours to train a model fully. This is quite a big difference, considering that the model must be trained for several epochs in order to achieve proper quality.

Another important result that was demonstrated in these experiments is that the AdaLoRA method is completely superior to the LoRA method on any dataset, with any basic model and metric. This happens mainly due to the algorithm, which allows to select the appropriate rank of the matrix for each layer and not set it fixed, as it happens in LoRA. This is a rather important result, which will be taken into account in the next section and when defining the final model. Another important fact is that AdaLoRA is superior to a fully fine-tuned model on some metrics. This was discussed in more detail in the second section, but this phenomenon may be explained by the fact that large networks after fine-tuning forget some of the information received during initial training, as a result losing quality. This confirmed the effect described in the original article on AdaLoRA.

\begin{figure}[htbp!]
\begin{center}
\includegraphics[width=0.95\columnwidth]{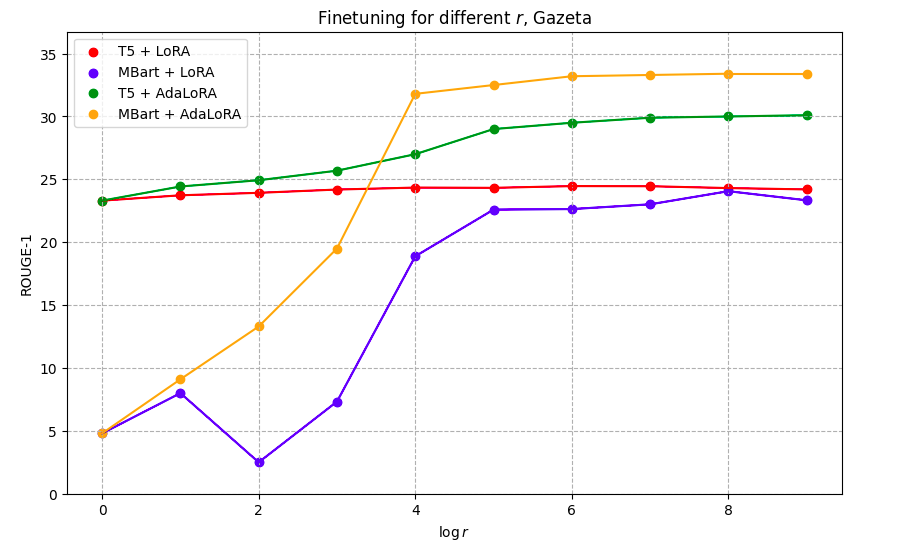}
\end{center}
\caption{ROUGE-1 metric for fine-tuning MBart and T5 models with adapters with different $r$ with Gazeta dataset}
\label{fig:gazeta_rouge}
\end{figure}

\begin{figure}[htbp!]
\begin{center}
\includegraphics[width=0.95\columnwidth]{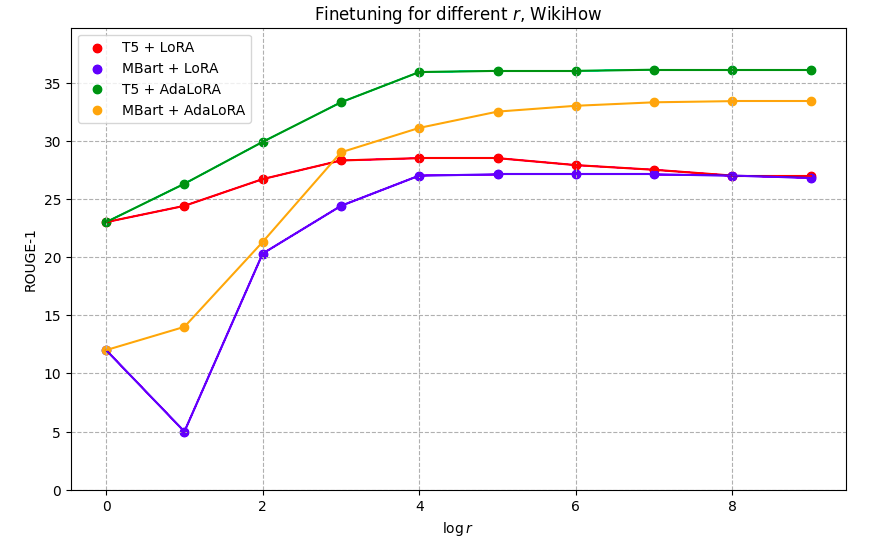}
\end{center}
\caption{ROUGE-1 metric for fine-tuning MBart and T5 models with adapters with different $r$ with WikiHow dataset}
\label{fig:wikihow_rouge}
\end{figure}

Now, as announced above, let's focus on changing metrics with different hyperparameter $r$. Below are graphs \ref{fig:gazeta_rouge} and \ref{fig:wikihow_rouge} which display the change in the ROUGE-1 metric for various $r$. Along the $x$ axis, the value of $r$ is represented in a logarithmic scale. For the rest of the metrics, graphs with similar trends are obtained, so they are not presented here.

The first thing that is noticeable on the charts is that for each model, at low $r$, the metric values for LoRA and AdaLoRA are very close. However, as the $r$ value increases, then ROUGE-1 metric increases too, which gets better for AdaLoRA. Another feature that is demonstrated by these graphs is that the optimal value of the metric is achieved with sufficiently small values of the hyperparameter $r$, roughly speaking in a diapason of 16-32. After that, the threshold value remains the same or gets slightly worse. Finally, it can be noted that for large values of $r$, the result of fine-tuning models with LoRA turns out to be independent of which model was taken as training. At the same time, for AdaLoRA, the best result is gained by the model that, even after fully training the scales, receives the best metric value.

\setcounter{page}{19}
\chapter{Automatic Speech Recognition}

\section{Whisper}

Whisper \cite{9} is a family of automatic speech recognition (ASR) models developed by OpenAI, representing a significant advancement in the field of speech recognition due to its robustness and versatility. These models are based on the Transformer architecture, a widely adopted model in natural language processing (NLP) tasks. The architecture consists of an encoder-decoder structure, where both the encoder and decoder are composed of multiple layers of self-attention and feed-forward neural networks. The encoder processes the input audio signal, extracting features and creating a representation that captures the essential information from the audio. The decoder then takes this encoded representation and generates the corresponding text transcription, involving attending to different parts of the encoded input and using context from previously generated tokens to predict the next token.

Whisper models come in various sizes, each differing in the number of layers and the size of the layers in the encoder and decoder. This results in models with different numbers of parameters, tailored to balance between computational efficiency and accuracy. For instance, Whisper-Tiny, the smallest model designed for low-latency applications, has approximately 39 million parameters. Whisper-Base is a slightly larger model with around 74 million parameters. Whisper-Small, an intermediate model, has around 244 million parameters. Whisper-Medium is a larger model with around 769 million parameters. The largest model, Whisper-Large, provides the highest accuracy with around 1.55 billion parameters. In the future, experiments will be conducted with each of these models and the results of the metrics will be compared.

These models were trained using a large and diverse dataset of audio recordings and their corresponding transcriptions. The dataset includes a wide variety of languages, accents, and acoustic conditions, which helps the model generalize well across different speech contexts. The training process involves converting audio signals into spectrograms, which are time-frequency representations of the audio serving as input to the model. Various data augmentation techniques are applied to artificially increase the diversity of the training data, such as adding background noise, varying the pitch, and changing the speed of the audio. The model is trained using gradient descent-based optimization techniques, such as Adam, to minimize the error between the predicted transcriptions and the ground truth.

Whisper models are known for their robustness in handling a wide range of speech variations, including different languages, accents, and dialects, thanks to their diverse training data. They also perform well in noisy environments, making them suitable for real-world applications where background noise is prevalent. These models can be applied to numerous tasks, including enhancing the accuracy and responsiveness of voice assistants like Siri, Alexa, and Google Assistant, providing accurate transcriptions for meetings, interviews, and media content, enabling real-time transcription and captioning services for the hearing impaired, and assisting language learners by providing accurate transcriptions and pronunciation feedback. In summary, Whisper models, with their advanced architecture, extensive training, and large parameter sizes, represent a state-of-the-art solution in the field of automatic speech recognition, offering high accuracy and versatility across various applications and languages.

\section{Dataset and metric}

It was not a very easy task to find a dataset for training the ASR model. While there are many datasets for training Audio2Text models, it was not enough for us only to have audio-text pairs, we also needed generated abstract summaries for each text. There are very few such datasets, and one of them is the How2 dataset \cite{37}. Although the reason for the need for brief contents will be discussed in the next section, however, it is worth noting this in order to understand the motivation for considering this particular dataset.

The How2 dataset is a rich collection of instructional videos, each paired with spoken utterances, English subtitles, crowdsourced Portuguese translations, and English video summaries. This dataset's extensive multimodality makes it an excellent resource for developing advanced models for multimodal understanding. Unlike other multimodal datasets, How2 features naturally occurring data: the subtitles and summaries are created by the original video creators and not crowdsourced, ensuring authentic and contextually relevant annotations. The visual content in the videos is inherently linked to the spoken utterances, offering additional features that can enhance model training. The dataset includes 79,114 instructional videos, totaling 2,000 hours of content, with an average video length of 90 seconds. These videos were sourced from YouTube and come with various types of metadata, including ground-truth subtitles and descriptions in English.

In addition to the loss function, which is used to train the model, it is necessary to determine the metric by which the quality of the models will be compared. Word Error Rate (WER) is a common metric used to evaluate the accuracy of ASR systems. It measures the difference between a reference transcription, which is the correct transcription, and the hypothesis transcription produced by the ASR system. WER is widely used because it provides a straightforward and interpretable way to quantify errors in speech recognition output. The formula for calculating WER is the sum of the number of substitutions, deletions, and insertions divided by the total number of words in the reference transcription. To compute WER, a dynamic programming algorithm similar to the Levenshtein distance is used to align the reference and hypothesis transcriptions, allowing the identification of substitutions, deletions, and insertions.

WER is crucial for comparing the performance of different ASR systems, with lower WER values indicating more accurate transcriptions. Despite its simplicity and widespread use, WER has some limitations. It treats all errors equally, regardless of context, does not account for syntactic or semantic correctness, and does not weight errors based on word significance. Nevertheless, WER is extensively used in benchmarking ASR systems, guiding improvements in ASR development, and academic research. It remains a fundamental metric for understanding and improving the performance of speech recognition technologies.

\section{Fine-tuning whisper model}

If during the training of the text summarization model, emphasis was placed on the study of various approaches to retraining AdaLoRA adapters, then at the stage of fine-tuning ASR models, the task of the current work was to investigate the influence of quantization described in section 2. The advantages of this approach are particularly valuable in that it not only reduces the fine-tuning time, but also allows you to significantly reduce the memory costs that the resulting model takes up.

Let's take a closer look at the experiments that have been done. I have done fine-tuning of Whisper models with various types of quantization (int4 and int8) with LoRA adapters, just with quantization and without quantization. At the same time, learning without quantization was done only for small models (tiny and small) due to the limitations of the computing power of the cluster. Below are the important hyperparameters that were used in the training for quantized models. This is quite important, since practice has shown that the values of the learning rate presented in the article are too large for fine-tuning, so practice has shown that it is required to take hyperparameters less than they are in the original article (proposed ones are 40 times less).

\begin{table}[ht]
\centering
\begin{tabular}{|| >{\centering\arraybackslash}m{3.6cm}| >{\centering\arraybackslash}p{4cm} | >{\centering\arraybackslash}p{4.5cm} | >{\centering\arraybackslash}p{2cm} ||} 
  \hline
       & learning rate (paper) & learning rate (proposed) & batch size \\ 
  \hline
  Tiny  (39M~parameters) & $1.5\times10^{-3}$  & $3.75\times10^{-5}$ & 8 \\
  \hline
  Base (74M~parameters) & $1\times10^{-3}$ & $2.5\times10^{-5}$ & 8 \\ 
  \hline
  Small (244M~parameters) & $5\times10^{-4}$ & $1.25\times10^{-5}$ & 4  \\ 
  \hline
  Medium (769M~parameters) & $2.5\times10^{-4}$ & $6.25\times10^{-6}$ & 2 \\ 
  \hline
  Large (1550M~parameters) & $1.75\times10^{-4}$ & $4.375\times10^{-6}$ & 1 \\ 
  \hline
\end{tabular}
\caption{Hyperparameters for fine-tuning quantized Whisper }
\label{tb:5}
\end{table}

Another important point is that the data contained in the dataset is supported in Kaldi format. This format of representation of sound features is not suitable for the Whisper model, so you can either remove the first layers of the encoder, or transform the features using the librosa library into mel-spectogram. Since the second method is simpler and does not require additional model training, it was chosen for this task.

Before moving on to quality metrics, let's talk about how much compression of models occurs, how many additional parameters in AdaLoRA adapters appear in the model and how long it takes to train one epoch of the model. These data are presented in the tables below. In addition to information about the absolute value of memory required to store model data, it can also be concluded how much memory is required to fully retrain these models. So in general, you can navigate that for full training of models, 10-15 times more GPU is required than the model weights require, and for inference, you need about 1.5 times more memory from what the model weights are trying. It is clear that these numbers strongly depend on the size of the batch, but these data are indicated on average.

\begin{table}[ht]
\centering
\begin{tabular}{|| >{\centering\arraybackslash}m{3.6cm}| >{\centering\arraybackslash}p{2.5cm} | >{\centering\arraybackslash}p{2cm} | >{\centering\arraybackslash}p{2cm} |>{\centering\arraybackslash}p{2.5cm} ||} 
  \hline
       & memory (initial model) & memory (int4) & memory (int8) & AdaLoRA compression \\ 
  \hline
  Tiny  (39M~parameters) & 75 MB  & 22MB & 40MB &1.0\%\\
  \hline
  Base (74M~parameters) & 142 MB & 42MB  & 76MB & 1.7\%\\ 
  \hline
  Small (244M~parameters) & 466 MB & 140MB & 265MB & 1.8\%\\ 
  \hline
  Medium (769M~parameters) & 1.5 GB & 460MB & 905MB & 2.0\% \\ 
  \hline
  Large (1550M~parameters) & 2.9 GB & 890MB & 1.7GB & 2.3\%\\ 
  \hline
\end{tabular}
\caption{Memory characteristics of Whisper}
\label{tb:6}
\end{table}

Speaking about the quality of compression, the table shows that with quantization in int4, the total volume of weights decreases by about 65 percent, and with quantization in int8, the total volume of weights decreases by 30 percent. At the same time, as it will be shown above, during further training, the quality of such models is not much inferior to the original model.

\begin{table}[ht]
\centering
\begin{tabular}{|| >{\centering\arraybackslash}m{3.6cm}| >{\centering\arraybackslash}p{2.5cm} | >{\centering\arraybackslash}p{2cm} | >{\centering\arraybackslash}p{2cm} |>{\centering\arraybackslash}p{2.5cm} ||} 
  \hline
       & initial model & int4 & int8 \\ 
  \hline
  Tiny  (39M~parameters) & 126min  & 38min & 53min \\
  \hline
  Base (74M~parameters) & 250min & 75min & 133min \\ 
  \hline
  Small (244M~parameters) & --- & 150min & 287min \\ 
  \hline
  Medium (769M~parameters) & --- & 445min & 823min \\ 
  \hline
  Large (1550M~parameters) & --- &  870min & 1530min\\ 
  \hline
\end{tabular}
\caption{Whisper time consuming for different approaches of fine-tuning per 1 epoch for ~40h of trainng data}
\label{tb:6}
\end{table}

\begin{figure}[!tbp]
  \includegraphics[width=0.49\textwidth]{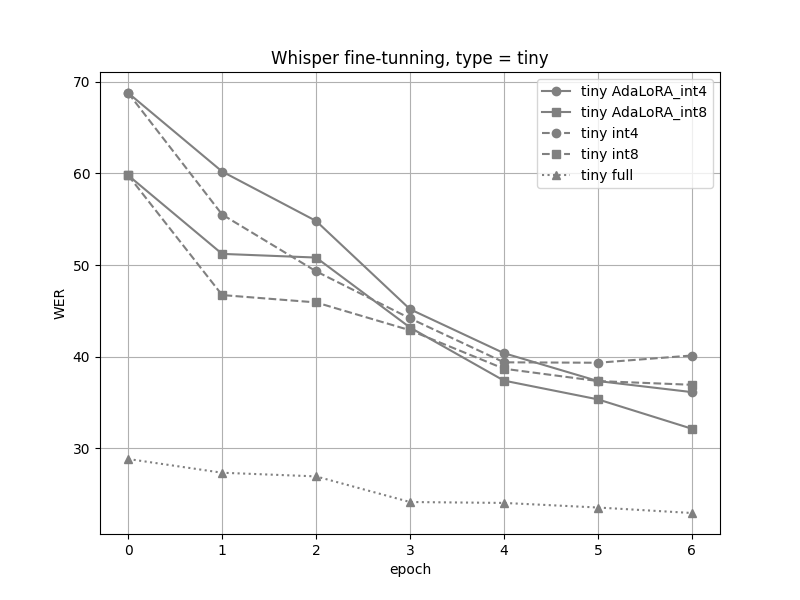} \hfill
  \includegraphics[width=0.49\textwidth]{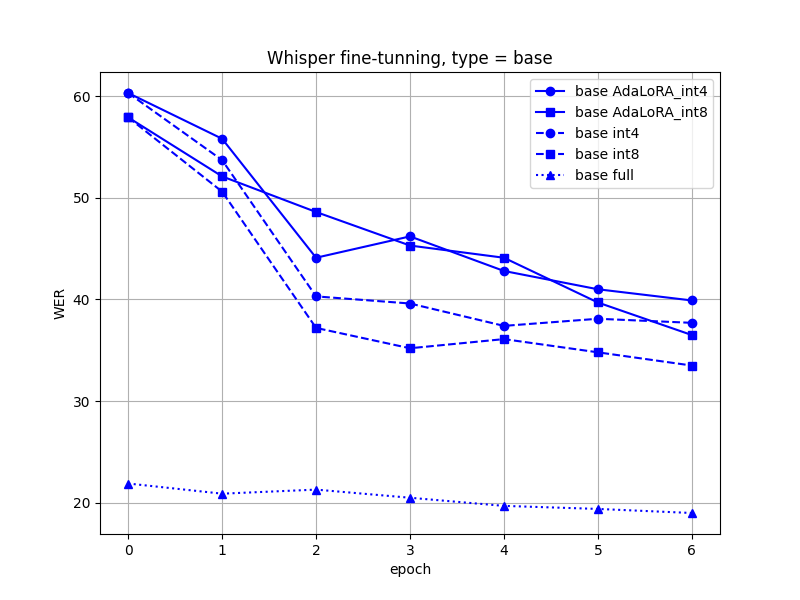}
  \\[\smallskipamount]

  \includegraphics[width=0.49\textwidth]{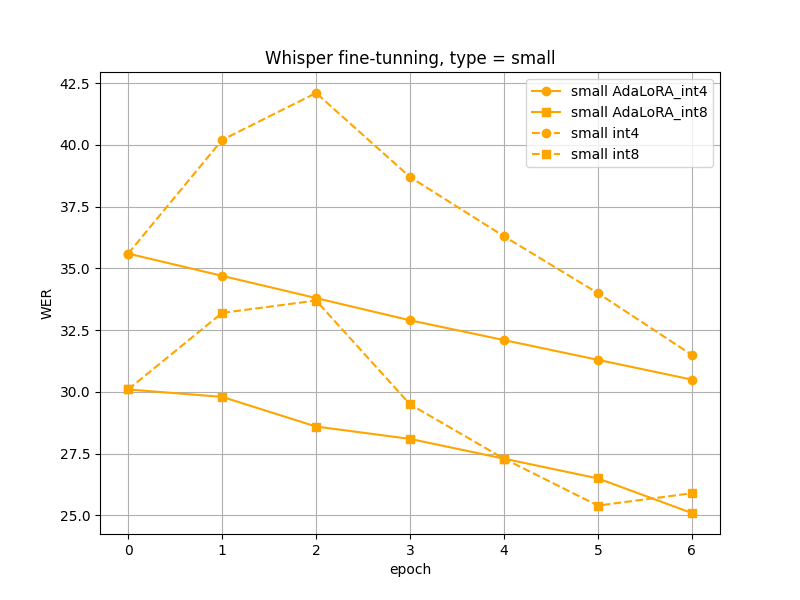}
  \hfill
  \includegraphics[width=0.49\textwidth]{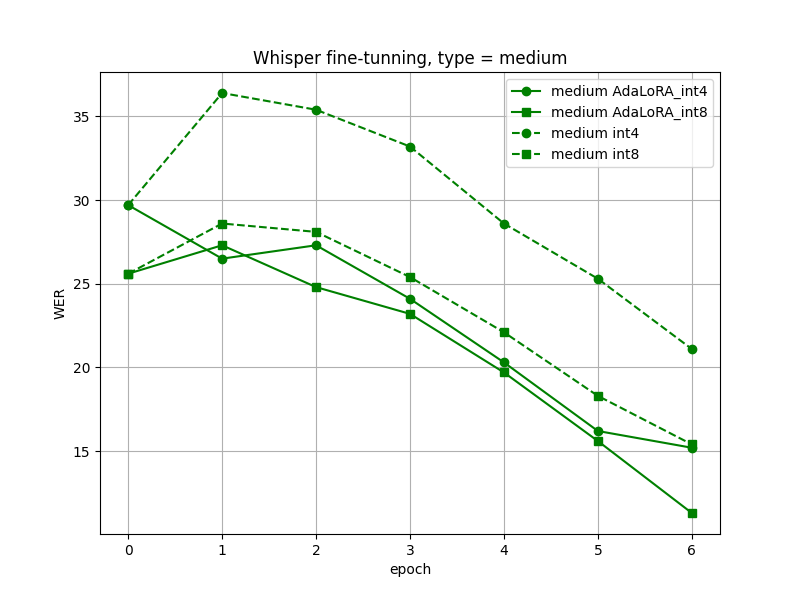}
  \\[\smallskipamount]
  \centering
  \includegraphics[width=0.49\textwidth]{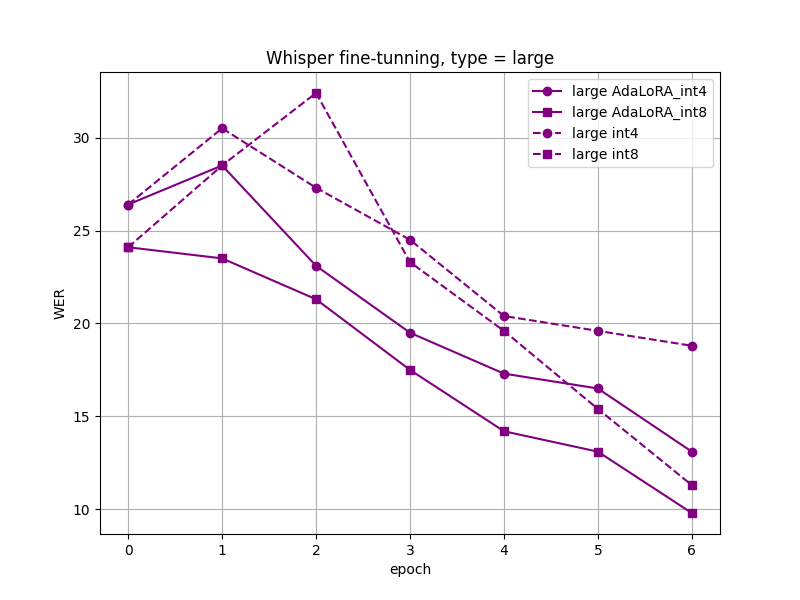}
  \caption{WER during fine-tuning Whisper model with different amount of weights}\label{fig:whisper1}
\end{figure}

\begin{figure}[!tbp]
  \includegraphics[width=0.49\textwidth]{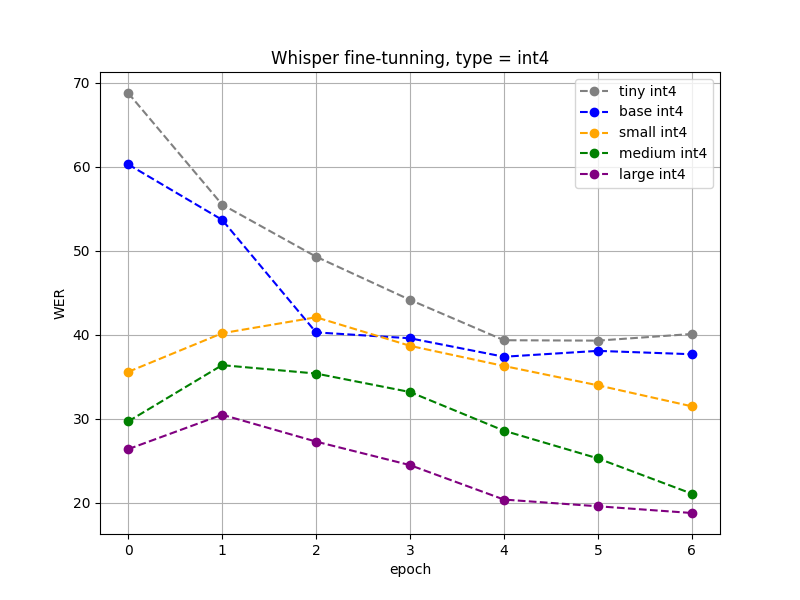} \hfill
  \includegraphics[width=0.49\textwidth]{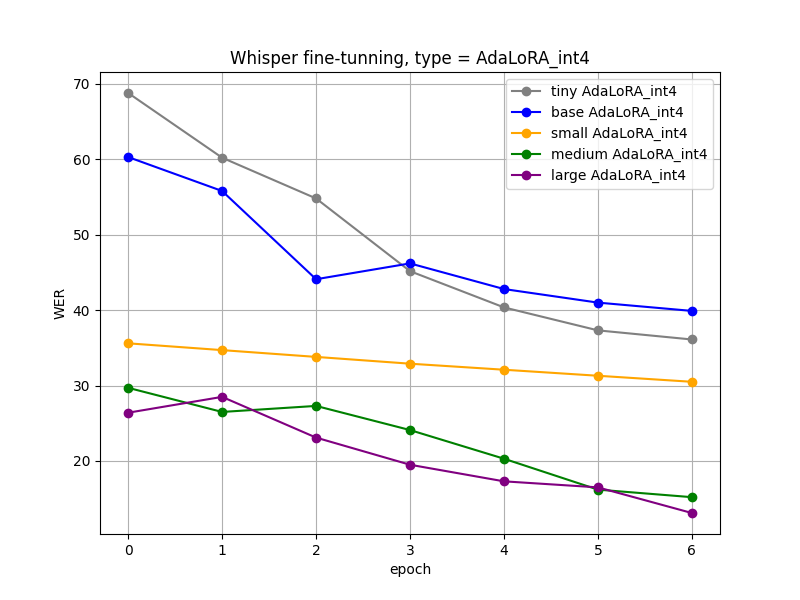}
  \\[\smallskipamount]

  \includegraphics[width=0.49\textwidth]{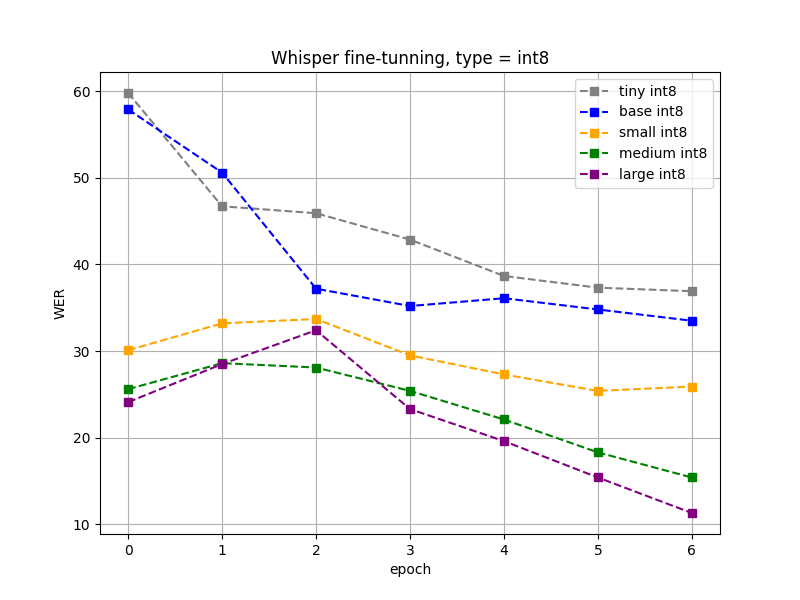}
  \hfill
  \includegraphics[width=0.49\textwidth]{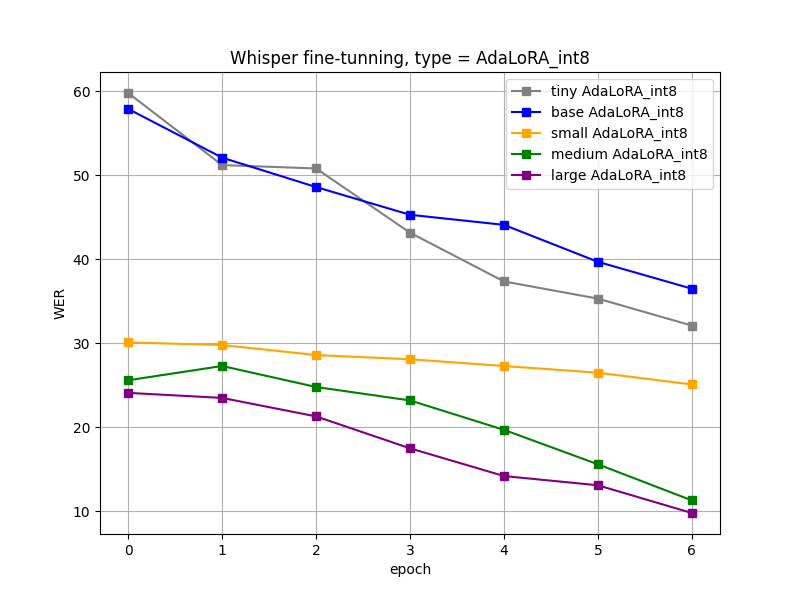}
  \\[\smallskipamount]
  \centering
  \includegraphics[width=0.49\textwidth]{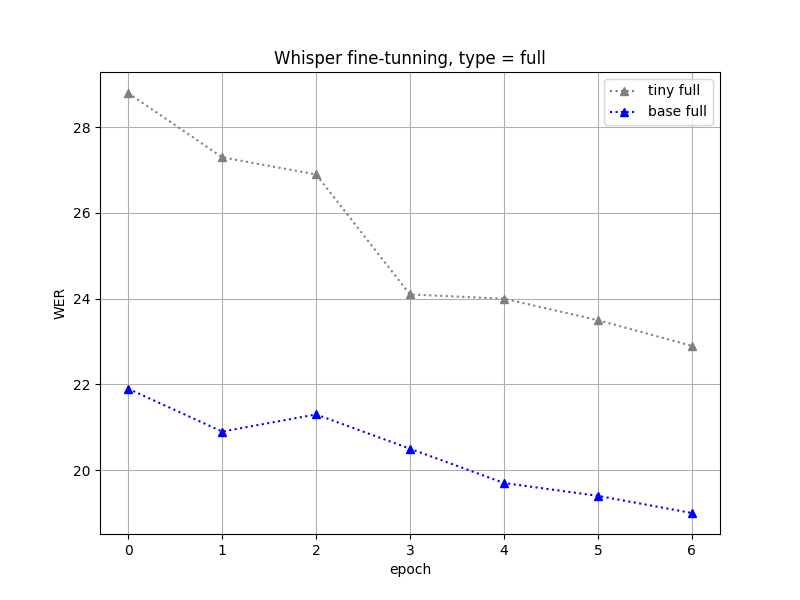}
  
  \caption{WER during fine-tuning Whisper model with different types of fine-tunning }\label{fig:whisper2}
\end{figure}

Now let's focus on the graphs of the WER metric for different models in the learning process for 6 epochs. Below are 10 graphs for each type of training (int4, int8, int4 + AdaLoRA, int8 + AdaLoRA) and each size of the Wisper model (tiny, base, small, medium, large). These 10 graphs are grouped according to two criteria. In Figure \ref{fig:whisper1}, each graph shows graphs of decreasing metrics for each of the five dimensions of the model. In Figure \ref{fig:whisper2}, each graph shows metrics for each type of training.

The following patterns can be found in graph \ref{fig:whisper1}. Firstly, it can be seen that the tiny and base models manage to learn in 6 epochs, while models based on small, medium, large retraining require additional additional training, the number of iterations of gradient descent in these neural networks is not enough. Another important pattern is that the WER graphs for the tiny and base models decrease immediately, starting from the first epoch. At the same time, this is not true for small, medium, large models: learning types without AdaLoRA adapters first worsen their metric, and only then increase it. This may be due to the fact that large networks, storing more information in themselves, lose some of the information obtained on the original dataset when the weights (even quantized ones) are fully fine-tuning. Moreover, such a large gap between two subsets of these two models may be due to a sharp increase in the number of weights during the transition from the base model to the small model (an increase in the number of weights by 3.3 times). A slightly less big difference is observed when switching from small to medium. However, there, too, as a possible consequence, there is a noticeable lack of training of models.

Let's focus on the Figure \ref{fig:whisper2}. A significant gap between two groups: tiny, base models and medium, large models can be spotted. It can be seen that within these two groups, the models give approximately the same result for each type of training, while the difference in the final metrics after 6 epochs is quite large with the small model. At the same time, the metric between the first group of models and the small model differs about as much as the metric between the second group and the small model. This may indicate a qualitative difference when switching to medium and large models. As a conclusion, in the further E2E model, it is necessary to focus specifically on the medium model, since it is noticeably better than other models with fewer weights. At the same time, its quality is not much inferior to the quality of the large model, but its number of weights is much higher (see the table \ref{tb:5}).

In addition, the graphs point in favor of quantization of the models. Indeed, learning without quantization is quite slow, only by a few percent of the WER metric. At the same time, even for small models, this training requires a lot of time. Of course, the graph shows a decreasing trend of the metric, which indicates that these models are untrained under the current dataset. However, as mentioned above, it is a difficult task to train them to the end. In addition, as can be seen from the graphs above, fine-tuned models using PEFT approaches show better results than fully fine-tuned models with a small number of parameters.

The table \ref{tb:7} below shows the best metric values after 6 training epochs for each model, for each type of training.

\begin{table}[ht]
\centering
\begin{tabular}{|| >{\centering\arraybackslash}m{3.6cm}| >{\centering\arraybackslash}p{2cm} | >{\centering\arraybackslash}p{1cm} | >{\centering\arraybackslash}p{1cm} |>{\centering\arraybackslash}p{1.7cm} |>{\centering\arraybackslash}p{1.7cm}||} 
  \hline
       & full fine-tuning & int4 & int8 & int4 AdaLoRA & int8  AdaLoRA\\ 
  \hline
  Tiny  (39M~parameters) & \textbf{22.9}  & 39.3 & 36.9 & 36.1 & 32.1 \\
  \hline
  Base (74M~parameters) & \textbf{19.0} &37.4& 33.5& 39.9& 36.5 \\ 
  \hline
  Small (244M~parameters) & --- & 31.5 & \textbf{25.4} &30.5 & \textbf{25.1}  \\ 
  \hline
  Medium (769M~parameters) & --- & 21.1 &15.4 &15.2 & \textbf{11.3}  \\ 
  \hline
  Large (1550M~parameters) & --- &  18.8 &11.3 &13.1 & \textbf{9.8} \\ 
  \hline
\end{tabular}
\caption{WER for different fine-tuned Whisper models after 6 epochs of training}
\label{tb:7}
\end{table}

In the table \ref{tb:7}, the best results within the same basic model are highlighted in each row. It can be seen that the int8 + AdaLoRA option is the best for large models.

\setcounter{page}{29}
\chapter{Conclusion and future work}

\section{Conclusion}

Our research presents a new effective approach for creating an E2E model of abstract audio summarization. When creating this model, various modern machine learning methods are used: large linguistic models and methods of effective fine-tuning of neural networks.

We investigated the LoRA and AdaLoRA learning methods on the task of text summarization. It was found out that, regardless of the basic model, the AdaLoRA method shows values better than LoRA, and values close to those given by a fully trained model. Thus, the effectiveness of these methods on the selected task was shown. Additionally, we investigated not only the compression parameter of the adapters with respect to the initial number of weights, but also found an optimal hyperparameter $r$ of the internal dimension of the adapter matrix in each of the methods.

On the other hand, in the course of our work, a family of ASR models was trained, which will later be used to build the final E2E audio summarization model. This training included not only obtaining good-quality models comparable to State-of-the-art models, but also researching the effects of quantification and AdaLoRA on the learning process. We compared the compression coefficients of the models, their training time, as well as the final quality obtained. In addition, hyperparameters were selected for successful training of these models. In addition, the effect of the learning quality of the model was investigated depending on the initial size of the weights. A qualitative difference was shown between the training of small and large models, expressed differently depending on the type of training.

At the end, a concept was proposed on how to make an E2E audio summation model from the above models, which should have a number of advantages compared to the already known version. On the one hand, such a model should have a lower total weight of the model, on the other hand, its training time should be less, and the process itself should require less computing resources.

\section{Future work}

For the remaining time of the internship, the main goal is to teach the final E2E model, which was proposed in this work. In addition, a fairly important goal is to check the results described in the article, which we were inspired by when conducting our work.

After that, we're going to focus on two areas of improving our model. On the one hand, we want to improve the compression of the model by investigating the influence of LoRA adapters, as was done in this work \cite{35}. In addition, an improvement in quality can be obtained with the addition of new features such as video frames of what is happening (a similar approach is used in this paper \cite{36}), or using a learning to rank approach \cite{39} when the S2T model uses abstract and extractive summarization approaches simultaneously. On the other hand, we want to improve the compression of the model. This can be done using new approaches, such as in this article \cite{34}, where other approaches are proposed to simultaneously use quantization and LoRA adapters.

\backmatter

\bibliographystyle{plain} 
\bibliography{bibfile.bbl}


\end{document}